\documentclass[conference]{IEEEtran}
\IEEEoverridecommandlockouts
\usepackage{cite}
\usepackage{amsmath,amssymb,amsfonts}
\usepackage{algorithmic}
\usepackage{graphicx}
\usepackage{textcomp}
\usepackage{xcolor}
\usepackage{array} 
\usepackage{tabularx} 
\usepackage{tabularray}

\newcolumntype{Y}{>{\centering\arraybackslash}X} 

\def\BibTeX{{\rm B\kern-.05em{\sc i\kern-.025em b}\kern-.08em
    T\kern-.1667em\lower.7ex\hbox{E}\kern-.125emX}}
\begin{document}

\title{DataAssist: A Machine Learning Approach to Data Cleaning and Preparation\\
}

\author{
\IEEEauthorblockN{Kartikay Goyle}
\IEEEauthorblockA{Department of Electrical and\\
Computer Engineering\\
University of Toronto\\
Toronto, Canada\\
kartikay.goyle@mail.utoronto.ca}\\   
\and
\IEEEauthorblockN{Quin Xie}
\IEEEauthorblockA{Department of Medical Biophysics \\
University of Toronto\\
Toronto, Canada\\
quin.xie@mail.utoronto.ca}
\and
\IEEEauthorblockN{Vakul Goyle}
\IEEEauthorblockA{School of Computer Science and Engineering \\
Nanyang Technological University\\
Singapore\\
vakul002@e.ntu.edu.sg}\\

}

\maketitle

\begin{abstract}
Current automated machine learning (ML) tools are model-centric, focusing on model selection and parameter optimization. However, the majority of the time in data analysis is devoted to data cleaning and wrangling, for which limited tools are available. Here we present DataAssist, an automated data preparation and cleaning platform that enhances dataset quality using ML-informed methods. We show that DataAssist provides a pipeline for exploratory data analysis and data cleaning, including generating visualization for user-selected variables, unifying data annotation, suggesting anomaly removal, and preprocessing data. The exported dataset can be readily integrated with other autoML tools or user-specified model for downstream analysis. Our data-centric tool is applicable to a variety of fields, including economics, business, and forecasting applications saving over 50\% time of the time spent on data cleansing and preparation.
\end{abstract}


\section{Introduction}
The rising availability of large datasets and computational power have enabled increased employment of multi-parameter, complex ML models and facilitated a wealth of autoML platforms that make ML models accessible for individuals with limited machine learning and programming expertise \cite{b1, b2, b3}. However, current autoML platforms do not provide support for the quality and integrity constraints of the dataset. Sub-optimal data quality may negatively impact ML model performance, though some models are insensitive to dirty data. Despite the joint cleaning data/ML problem being recognized by both ML and database (DB) communities, there is no standardized solution in a common workflow for a data analyst: They perform exploratory data analysis (EDA) by plotting variable distributions and inspecting trends between variables to determine potential predictors of the response variable. They input a subset of the data into a preliminary ML model and discover idiosyncrasies in the dataset during training or earlier EDA. They would need to manually clean the data or implement some data-cleaning libraries, and retrain the model iteratively until there is no dirty data.
\\

The iterative nature of the above procedure makes data preprocessing unnecessarily time-consuming and repetitive, taking time away from the ultimate goal of making interpretations and deriving knowledge. A centralized tool for principled data cleaning will effectively free data scientists from the laborious process of data preparation, integration, and management. Over the years, the database community has developed different types of data-cleaning tools but the data-cleaning ecosystem remains diffuse. The analytics-driven data cleaning tools reduce the cost of data cleaning by simultaneously integrating data cleaning and training. For example, ActiveClean \cite{b4} takes advantage of the gradient descent method, allowing the cleaning of small batches of data on ML models with convex loss, such as linear regression and mixture models. BoostClean\cite{b5} addresses a selected set of errors with statistical boosting, but only considers when an attribute value is outside of its value domain. Some major disadvantages of these methods are repeated efforts in cleaning datasets when fitting different ML models, and difficulties in comparing across model performance.
\\

Another paradigm of data cleaning tools takes advantage of ML methods in correcting data annotation. Scared \cite{b6}, GDR\cite{b7}  and HoloClean \cite{b8} use different ML models to come up with probabilities for detected errors and suggest repair. In addition, active learning \cite{b9} has been used to prompt the users to label specific data potentially erroneous and receive timely feedback to prevent error propagation in downstream analysis. Nonetheless, there is no streamlined process to prepare the data for ML models. The data cleaning system is decentralized, with each tool specialized for a limited subset of error detection and repair.
In this paper, we present DataAssist, an ML-based pipeline integrating data exploration and cleaning that allows users to select, combine, and order different steps that best suit their data analysis scheme. The package covers the four most common sources of error in dirty data: missing values, outliers, duplicates, and inconsistencies. In addition, DataAssist provides data exploration and transformation functions, such as enumerating variable types, generating data visualization, ranking feature importance, adjusting data skewness, encoding categorical variables, and normalization. We designed a user interface that allows individuals with minimal to no coding experience to easily employ our tool and export datasets for downstream analysis.

\section{Feature Overview}
Our goal is to develop a standardized pipeline to automatically clean and prepare the data such that the users can specify their needs and directly integrate the data cleaning process with the existing autoML tools for the next steps such as feature engineering and model selection. We leveraged ML models to predict the most suitable methods for Exploratory data analysis (EDA) and preprocessing for each dataset.

\begin{figure*}[htbp]
\centering
\includegraphics[width=1\linewidth]{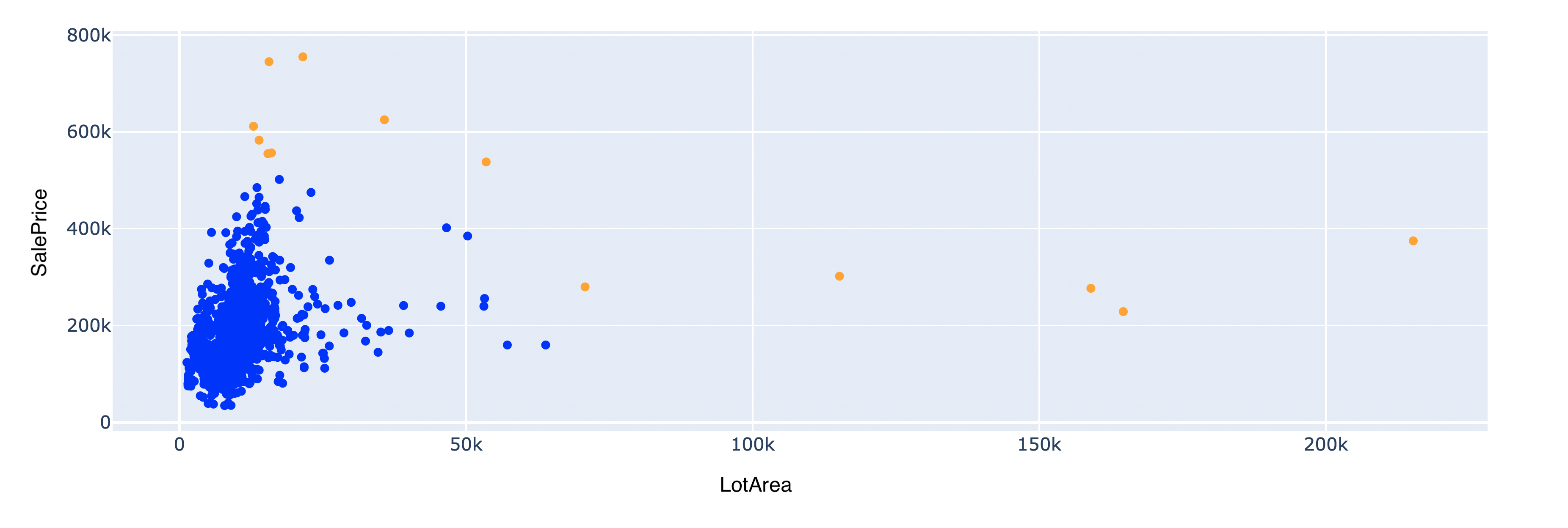}
\caption{Example output of anomaly detection from DataAssist. The plot is generated by running DataAssist on the House Prices dataset introduced in Section III. The x-axis is Lot size in square feet (LotArea) and the y-axis is the SalePrice of residential homes. The outliers are detected by DBSCAN and highlighted in orange, distinguished from the majority of the dataset in blue. These points will be removed in the subsequent data-cleaning steps.}
\label{fig1}
\end{figure*}

\renewcommand{\arraystretch}{1.5} 
\begin{table*}[ht]
\centering
\caption{EXAMPLE TRAINING DATASET FOR THE SVM MODEL PREDICTING EDA}
\label{table:1}
\begin{tabularx}{\textwidth}{|>{\centering\arraybackslash}X|>{\centering\arraybackslash}X|>{\centering\arraybackslash}X|>{\centering\arraybackslash}X|>{\centering\arraybackslash}X|>{\centering\arraybackslash}X|>{\centering\arraybackslash}X|>{\centering\arraybackslash}X|>{\centering\arraybackslash}X|>{\centering\arraybackslash}X|>{\centering\arraybackslash}X|}
\hline
\textbf{Variable 1 Name} & \textbf{Variable 1 Type} & \textbf{Variable 1 Distribution} & \textbf{Variable 2 Name} & \textbf{Variable 2 Type} & \textbf{Variable 2 Distribution} & \textbf{Plot Type} & \textbf{Variable Relation} & \textbf{Correlation Coefficient} \\
\hline
Age & Continuous & Normal & Income & Continuous & Skewed Right & Scatter Plot & Positive Linear & High Positive (~.8) \\
\hline
Gender & Categorical & Equal Male/Female & Purchase & Categorical & Varied & Bar Chart & No Relation & Not Applicable \\
\hline
Product\_type & Categorical & Varied & Sales & Continuous & Normal & Violin Plot & Positive Relation & Not Applicable \\
\hline
Experience & Continuous & Normal & Salary & Continuous & Skewed Right & Scatter Plot & Positive Linear & High Positive (~0.7) \\
\hline
Education & Ordinal & Varied & Genders & Categorical & Equal Male/Female & Bar Chart & No Relation & Not Applicable \\
\hline
Skill\_Level & Continuous & Normal & Task\_Time & Continuous & Normal & Line Graph & Negative Linear & High Negative \\
\hline
Fruit\_Type & Categorical & Varied & Popularity & Continuous & Varied & Pie Chart & No Clear Relation & Not Applicable \\
\hline
City & Categorical & Varied & Population & Continuous & Skewed Right & Bar Chart & Positive Relation & Not Applicable \\
\hline
\end{tabularx}
\end{table*}

\subsection{Exploratory data analysis}

First, EDA is the primary task of data analysis for research questions designed to hypothesis-generating rather than hypothesis-driven. Moreover, by inspecting the dataset during EDA, one may find idiosyncrasies in the dataset, such as missing values and outliers, skewed distribution, or multi-collinearity across variables. Therefore, we placed EDA at the forefront of DataAssist pipeline. Users can review data quality through a number of visualizations:
\\

\begin{itemize}

\item \textbf{User-selected variables.} The most suitable plots for variables are predicted by DataAssist based on rule-based systems and SVM models. The training dataset for the SVM model is a curated dataset based on hundreds of Kaggle notebooks, which can in turn be broken down into records of how users analyzed their datasets and the datasets themselves. The information included in our training dataset (Table~\ref{table:1}) are Variable 1 Name (Age, Income), Variable 1 Type (Categorical, Continuous, Ordinal), Distribution of Variable 1 (Normal, Skewed left), Variable 2 Name (Salary, Gender), Variable 2 Type (Categorical, Continuous, Ordinal), Distribution of Variable 2 (Normal, Skewed right), Variables Relation (Positive, Negative, Zero), Correlation Coefficient Threshold, Covariance threshold (Positive), Chi-square/ANOVA P-value Theoretical (High, Low), Mode/Cause of Relationship and the target variable plot type. As an output, plots that are most appropriate for the variable will be determined by our SVM model based on the type (categorical vs numerical) and distribution of the variable. The resulting plots, histograms, scatter plots, bar plots, box plots, violin plots, and/or alluvial plots, are standard procedures of EDA and will facilitate data quality control, model choice, and feature selection. In addition, the relationship between variables will be presented as heatmaps or cluster plots through analyses such as covariance/correlation analysis, hierarchical clustering and K means clustering.
\\
\item \textbf{Automatic feature selection.} At a holistic level, DataAssist performs association rule mining on the dataset via ML algorithms such as Apriori \cite{b10} and FP-growth \cite{b11} to discover related variables, for example, frequent itemset conforming to certain transaction rules. Moreover, DataAssist uses random forests and decision trees to rank important features.

\end{itemize}

\subsection{Data cleaning}

Subsequently, DataAssist offers the data preprocessing framework for data cleaning and preparation. We formulate the problem of data cleaning as follows: Given a set of constraints $L$, a structured dataset $D$ where $D \nvDash L$, and a dataset reflecting the desired data distribution $D_I$, identify and repair erroneous records such that the repair dataset $D_r \vDash L$ and the distance between $D_r$ and $D_I$ is minimized. $D$ is characterized by a set of attributes $A = \{A_1, A_2, ..., A_N\}$ which are essentially columns of the dataset. $D$ can also be represented as a set of rows, or vectors $V = \{v_1, v_2, ..., v_M\}$ where each $v_i$ consists of a set of cells denoted as $c[v_i] = \{A_i[v_i]\}$.  $v_i[A_j]$ the $j$-th cell of $i$-th vector for attribute $A_i \in A$. Users are allowed to select and specify the following features:
\\

\begin{itemize}

\item \textbf{Missing Values.} Missing values are detected by empty entries or infinity or NaNs. To avoid causing errors or overflow in ML models, users may choose to remove that row from the dataset if the number of missing values in a row exceeds a certain threshold. Additionally, the SVM model underlying DataAssist can decide the most suitable missing imputation technique by analyzing the variable in the data and the broader attribute distribution. Specifically, statistics like mean and median can be used to impute numerical variables and mode for categorical variables. Additionally, for both types of variables, the SVM model may recommend imputing the missing value through Multivariate Imputation by Chained Equations \cite{b12}, which iteratively runs ML models on other available features to predict the missing record.
\\

\item \textbf{Outliers.} Outliers. DataAssist offers a variety of anomaly detection algorithms $C(.)$: $A \rightarrow C \subseteq D$ for users to choose from. Univariate outliers $v_i[A_j]$ are detected by algorithms, such as IQR, Density-based spatial clustering of applications with noise (DBSCAN) \cite{b13} and isolation forest \cite{b14} if $v_i[A_j] \notin \{c_1, c_2... c_K | c_i \notin C\}$ ; multivariate outliers $v_i[A_j, A_k, A_l]$  are detected by local outlier factor. Upon completion of the pipeline, users are prompted to visualize the outliers highlighted by a different color from the rest of the dataset (Fig.~\ref{fig1}), and remove them by clicking on the outliers. Alternatively, users may choose to replace outliers with another value through winsorization.
\\

\item \textbf{Duplicates and inconsistencies.} Duplicates refer to records referring to the same real-world entity, whereas inconsistencies are the same entity with different representations. DataAssist employs a learnable similarity function $S(.) :V \rightarrow R$ in the form of pairwise supervision which consists of object pairs known to be similar or dissimilar \cite{b15}. $v_i[A_j]$  is substituted by $v_h[A_j]$ if $S(v_i[A_j], v_h[A_j]) \leq r_1$, a threshold for dissimilarity within attribute $A_j$. Alternatively, $v_j$ is removed for $S(v_i , v_h) \leq r_n$, a threshold that sets the minimum dissimilarity across attributes $A_j$'s.
\\

\end{itemize}

\renewcommand{\arraystretch}{1.5}
\begin{table*}[ht]
\centering
\caption{EXAMPLE TRAINING DATASET FOR THE XGBOOST MODEL PREDICTING PREPROCESSING.}
\label{table:2}
\begin{tabularx}{\textwidth}{|>{\centering\arraybackslash}X|>{\centering\arraybackslash}X|>{\centering\arraybackslash}X|>{\centering\arraybackslash}X|>{\centering\arraybackslash}X|>{\centering\arraybackslash}X|>{\centering\arraybackslash}X|>{\centering\arraybackslash}X|>{\centering\arraybackslash}X|>{\centering\arraybackslash}X|>{\centering\arraybackslash}X|>{\centering\arraybackslash}X|}
\hline
\textbf{Variable Name} & \textbf{Original. Distribution} & \textbf{Missing Value Handling} & \textbf{Transformat-ion} & \textbf{Feature Scaling} & \textbf{New Distribution} & \textbf{Required Analysis Type} & \textbf{Consider Outlier Treatment} & \textbf{Variable Nature} & \textbf{Scale of Measurement} \\
\hline
City & Categorical (100+ categories) & None & Frequency Encoding & None & Varied (fewer categories) & Any & No & Predictor & Nominal \\
\hline
Gender & Categorical ('Male', 'Female') & None & Label Encoding & None & Binary ('0', '1') & Any & No & Predictor & Nominal \\
\hline
Income & Continuous (wide-range) & Median Imputation & None & Min-Max Scaling & Continuous & Yes & Target & Ratio & High \\
\hline
Age & Continuous & None & Discretization (Age Groups) & None & Categorical & Classification & No & Predictor & Ratio \\
\hline
Product\_Type & Categorical (many categories) & Mode Imputation & One-Hot Encoding & None & Multiple-Binary & Any & No & Predictor & Nominal \\
\hline
Experience & Left Skewed & Mean Imputation & Square transformation & None & Approximate Normal & Any & Yes & Predictor & Ratio \\
\hline
Salary & Very wide range & None & None & Min-Max Normalization & Range between 0-1 & Regression & Yes & Target & Ratio \\
\hline
Height & Slightly left skewed & None & None & Z-score standardization & Normal Distribution & Any & Yes & Predictor & Ratio \\
\hline
\end{tabularx}
\end{table*}

\subsection{Data preprocessing}
Lastly, DataAssist offers preprocessing options such as data transformation to allow variables to conform to distributional assumptions required by ML models. We leveraged Gradient Boosting Machines like XGBoost \cite{b16} to train a multi-class multi-label classification task to predict preprocessing steps as target variables. The training dataset (Table~\ref{table:2}), similar to the one used in the previous model for predicting EDA, consists of information derived from the Kaggle dataset, including the original distribution (Categorical, or Continuous), Missing Value Handling (Mean, Median, Mode Imputation), Transformation (Frequency Encoding, Label Encoding, One-Hot encoding) , Feature Scaling (Min-Max Scaling, Z-Score standardization), New distribution, the variable type after transformation (Categorical, Binary), the type of analysis being performed (Classification, Regression), the target variable, the predictor variables, the scale of measurement (Nominal, Ordinal, Ratio), the Variable cardinality (Low, Med, High) and the Data Scope (Transactional, Demographic). Moreover, DataAssist takes advantage of the natural language processing (NLP) model BERT \cite{b17} to embed attributes $A$ of the same variable type as vectors in high dimensional spaces and uses different metrics of geometric distance [need >1] between $A_i$'s to determine their semantic similarity. If $dist(A_i, A_j) \leq d_n$, the distance at which similarity between $A_i$ and $A_j$ is calculated to be high by DataAssist, the program will perform the same data manipulation on these attributes.
\begin{itemize}

\item \textbf{One-hot/Label Encoding} 
It is important for algorithms to distinguish between the ordinal categorical variables for which attributes of significance correspond to the order of numbers from the nominal categorical variables for which orders do not matter. Label encoding uses, as a default option, integers to represent ordinal categorical variables. In contrast, for nominal categorical variables, a pre-processing step that converts an $A_c$ into a binary vector $A_c'$ whose entry is an indicator for whether a particular category is present ($A_c' = I(A_c = k) \forall k \in domain (A_c)$) is used to integrate them into ML models.
\\

\item \textbf{Standardization/Normalization} 
One of the most common ways of standardization is to center the numeric variable $A_n$ around 0 mean and scale it to unit variance such that $A_n \sim P (\mu = 0, \sigma = 1)$, where $P$ is any probability distribution, but most commonly a normal distribution, also known as z score normalization. Alternatively, it is possible to perform min-max normalization such that $A_n \in [0, 1] | [-1, 1]$ This is beneficial for ML algorithms that are sensitive to the scale of the features, such as support vector machines (SVMs) and k-nearest neighbors (KNN).
\\

\item \textbf{Power Transformation} 
This is another common pre-processing for numeric variables by applying power functions $A_n^\lambda$ where $\lambda$ is the power parameter, in order to make the distribution closer to normal distribution. DataAssist uses Box-cox transformation to determine the value of $\lambda$ through the likelihood function and goodness-of-fit tests. Some special cases of $\lambda$ include square root transformation, log transformation and log10 transformation.
\\
\end{itemize}

\begin{figure}[htbp]
\centering
\includegraphics[width=0.45\textwidth]{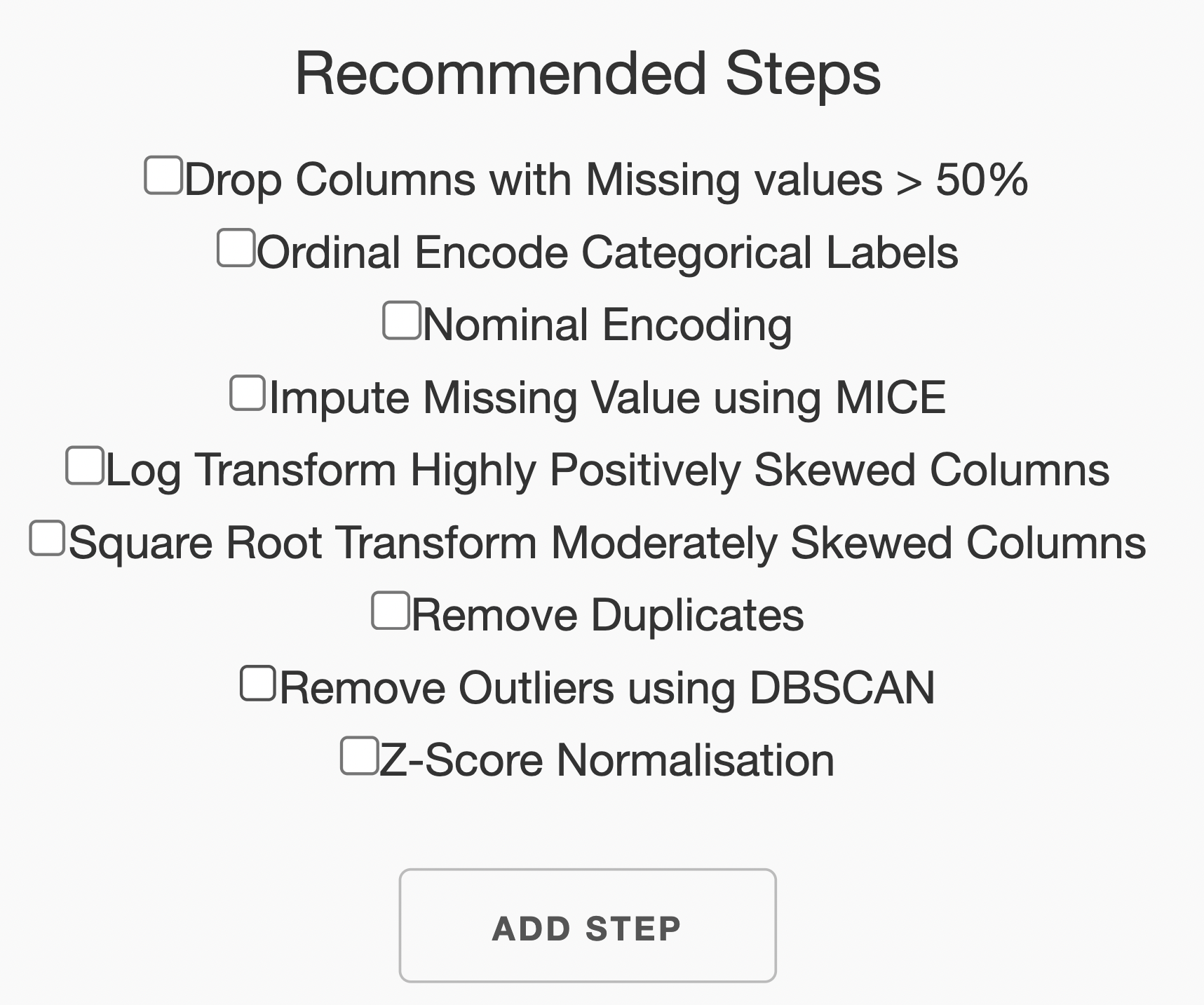}
\caption{Example of a figure caption.}
\label{fig3}
\end{figure}

\section{The Interface}
This section will use two datasets as an example to demonstrate the components of DataAssist. The user first navigates to the “Gather Data” page to preview and upload the dataset. Once uploaded, DataAssist will automatically perform the pipeline on the dataset.

\subsection{House prices}
This dataset \cite{b18} contains 1460 records of residential homes in Ames, Iowa sold between 2006 and 2010. Each record has 80 attributes including sale price, overall condition of the house, type of dwelling, and proximity to various conditions. The regression task is to predict the sale price using all or some of the attributes.

\begin{figure}[htbp]
\centering
\includegraphics[width=0.5\textwidth]{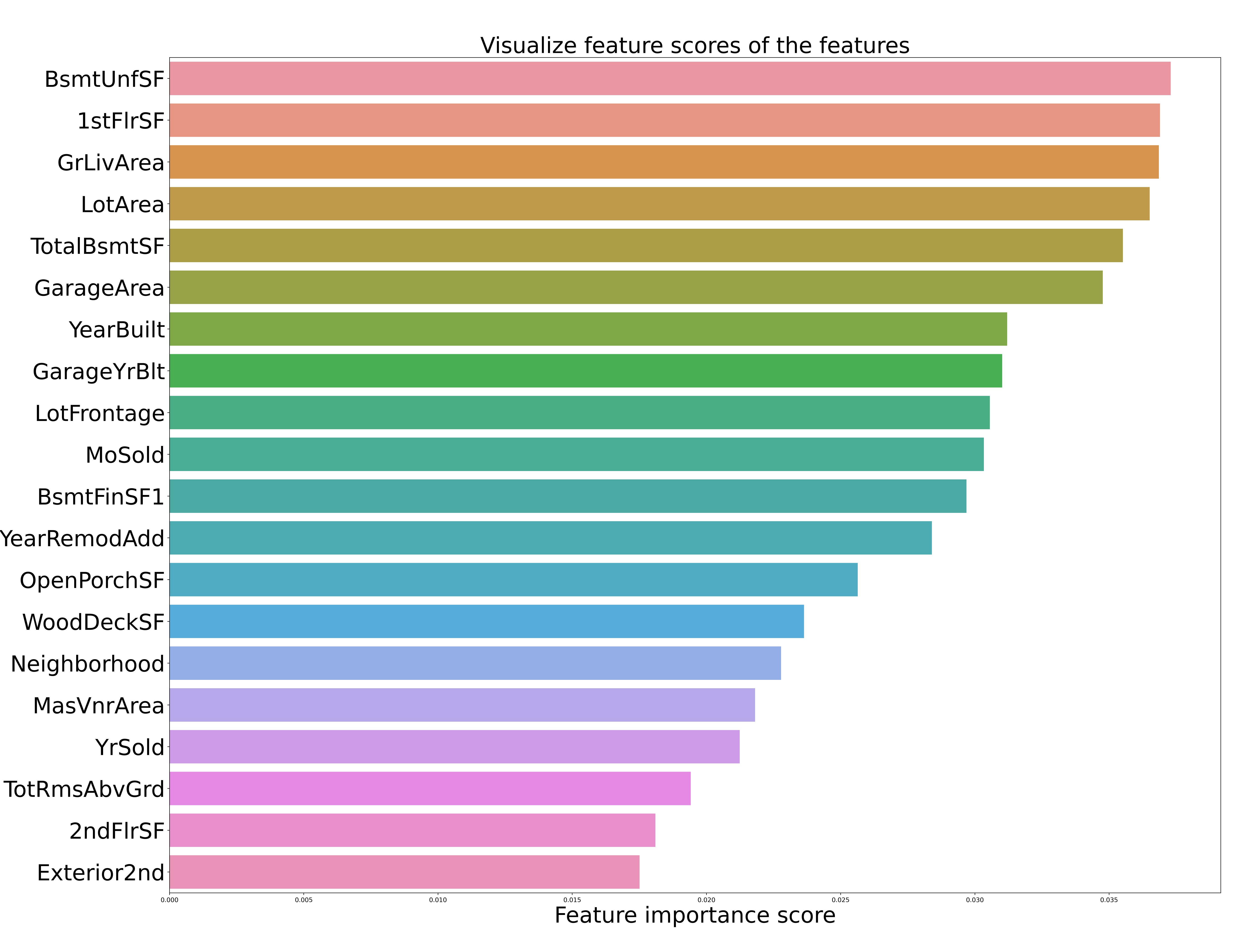}
\caption{Ranking of important Features associated with Sale Price.}
\label{fig2}
\end{figure}

As a part of EDA, DataAssist ranks the importance of the feature using random forest, shown in Fig.~\ref{fig2}. The top three important features associated with Sale Price are Unfinished square feet of basement area (BsmtUnfSF), First Floor square feet (1stFlrSF), and Above grade (ground) living area square feet (GrLivArea). Users are invited to inspect the distribution of specific higher-ranked important variables or select features based on external information based on domain knowledge. DataAssist then recommends steps for data cleaning and preparation, shown in Fig.~\ref{fig3}. Briefly, 38 records are removed from the dataset due to missing values. Multiple attributes, such as zoning classification of the sale (MSZoning), physical locations within Ames city (Neighborhood), and proximity to various conditions (Condition1, Condition2) are one-hot encoded. Categorial variables such as the overall condition of the house (OverallCond), the overall material and finish of the house (OverallQual), and Heating quality and condition (HeatingQC) are nominal encoded. The outliers shown in Fig.~\ref{fig1} are removed from the dataset. The numerical values are transformed by applying power functions, and subsequently normalized.

\subsection{Air Quality Data in India}\label{AA}
This dataset \cite{b19} contains 29531 records of daily air quality data across 26 cities in India. Each record has 15 attributes including datetime of the collection, PM 2.5, PM10, and concentration of various pollution gas. The regression or classification task is to rate the air quality index given the attributes.

As a part of EDA, DataAssist generates a heatmap with columns ordered by hierarchical clustering to explore the relationships between variables. Again, users are invited to inspect the distribution of variables of interest. If, for example, the daily PM2.5 level and City are selected, DataAssist will generate a barplot for the number of records corresponding to each city and a histogram for PM2.5. To demonstrate the relationship between two selected variables, DataAssist produces a violin plot showcasing the daily PM2.5 level for each city in India. For data cleaning and preprocessing, DataAssist recommends steps similar to ones shown in Fig.~\ref{fig3}. Briefly, 3582 records are removed from the dataset due to missing values. The only two categorical variables, City and AQL\_Bucket, are one-hot encoded and nominal encoded, respectively. The numerical values for PM2.5 and PM10 values, as well as various pollution gas concentrations, are standardized and normalized.

\section{DISCUSSION}

Numerous data-cleaning libraries have been developed over the years in an effort to integrate data cleaning and data analysis, a need increasingly recognized by both the DB community and the ML community. However, these tools focus on specific sets of error detection and repair, behooving the users to identify tools most suited to their needs from different sources and build their own pipelines. We address this problem with DataAssist, a centralized platform that integrates exploratory analysis, data cleaning, and pre-processing. DataAssist is an innovative application of ML models on the problem of, figuratively, meta-data-analysis. At various stages of the data preparation pipeline handled by DataAssist, several ML models were combined and built with engineered features to optimize the prediction task of the most appropriate method for data analysis, or more precisely, various steps of data manipulation that conform the input data to the assumptions or requirements of different ML models. We demonstrated the features of DataAssist using two public datasets, but the application of DataAssist is not limited to these fields. Wherever data are imperfectly curated, DataAssist can be used to automatically perform EDA, data cleaning, and preprocessing. In the future, DataAssist will be integrated with autoML tools to fully automate the process from data collection and cleaning to model training and deployment, further accelerating the process of deriving knowledge from real world data. We anticipate that DataAssist will broadly benefit individuals specialized in different fields, including economics, business, and forecasting applications.

\vspace{12pt}
\end{document}